\documentclass{article}

    \PassOptionsToPackage{numbers, compress}{natbib}

    \usepackage[preprint]{neurips_2023}

\usepackage[utf8]{inputenc} 
\usepackage[T1]{fontenc}    
\usepackage{hyperref}       
\usepackage{url}            
\usepackage{booktabs}       
\usepackage{amsfonts}       
\usepackage{nicefrac}       
\usepackage{microtype}      
\usepackage{xcolor}         

\usepackage{amsmath, amssymb, amsthm}
\usepackage{wrapfig}
\usepackage{xspace}

\usepackage{scrextend}

\usepackage{graphicx}

\usepackage{mathrsfs}
\usepackage{subcaption}

\usepackage{scalerel,graphicx,xparse}
\usepackage{hyperref}

\NewDocumentCommand\modelname{}{\scalerel*{\includegraphics{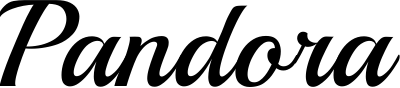}}{X}\xspace}

\NewDocumentCommand\icon{}{\raisebox{-0.1cm}{\includegraphics[width=0.7cm, height=0.637cm]{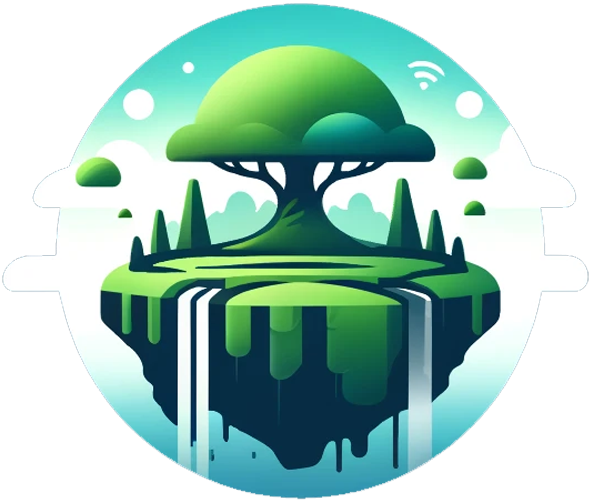}}}

\NewDocumentCommand{\gy}{ mO{} }{\textcolor{cyan}{\textsuperscript{\textit{Yi}}\textsf{\textbf{\small[#1]}}}}
\NewDocumentCommand{\bert}{ mO{} }{\textcolor{green}{\textsuperscript{\textit{Bert}}\textsf{\textbf{\small[#1]}}}}

\title{\icon\modelname: Towards General World Model \\ with Natural Language Actions and Video States}

\author{%
\small
Jiannan Xiang$^*$,~
Guangyi Liu$^*$,~
Yi Gu$^*$,~
Qiyue Gao,~
Yuting Ning,~
Yuheng Zha,\\
\small
\textbf{
Zeyu Feng,~
Tianhua Tao,~
Shibo Hao,~
Yemin Shi,~
Zhengzhong Liu,
}\\
\small
\textbf{Eric P. Xing,~~
Zhiting Hu}\\
{\small Maitrix.org,~ UC San Diego,~ MBZUAI~~~~ }
\\ 
\small\url{https://world-model.ai}~~~~ {\it *equal contribution}}

\begin{document}

\maketitle


\begin{figure}[!h]
\vspace{-28pt}
  \centering
  \caption{\modelname simulates future world states (videos) under action control (natural language).}
  \vspace{-5pt}
  \label{fig:examples}
  \includegraphics[width=1.01\textwidth]{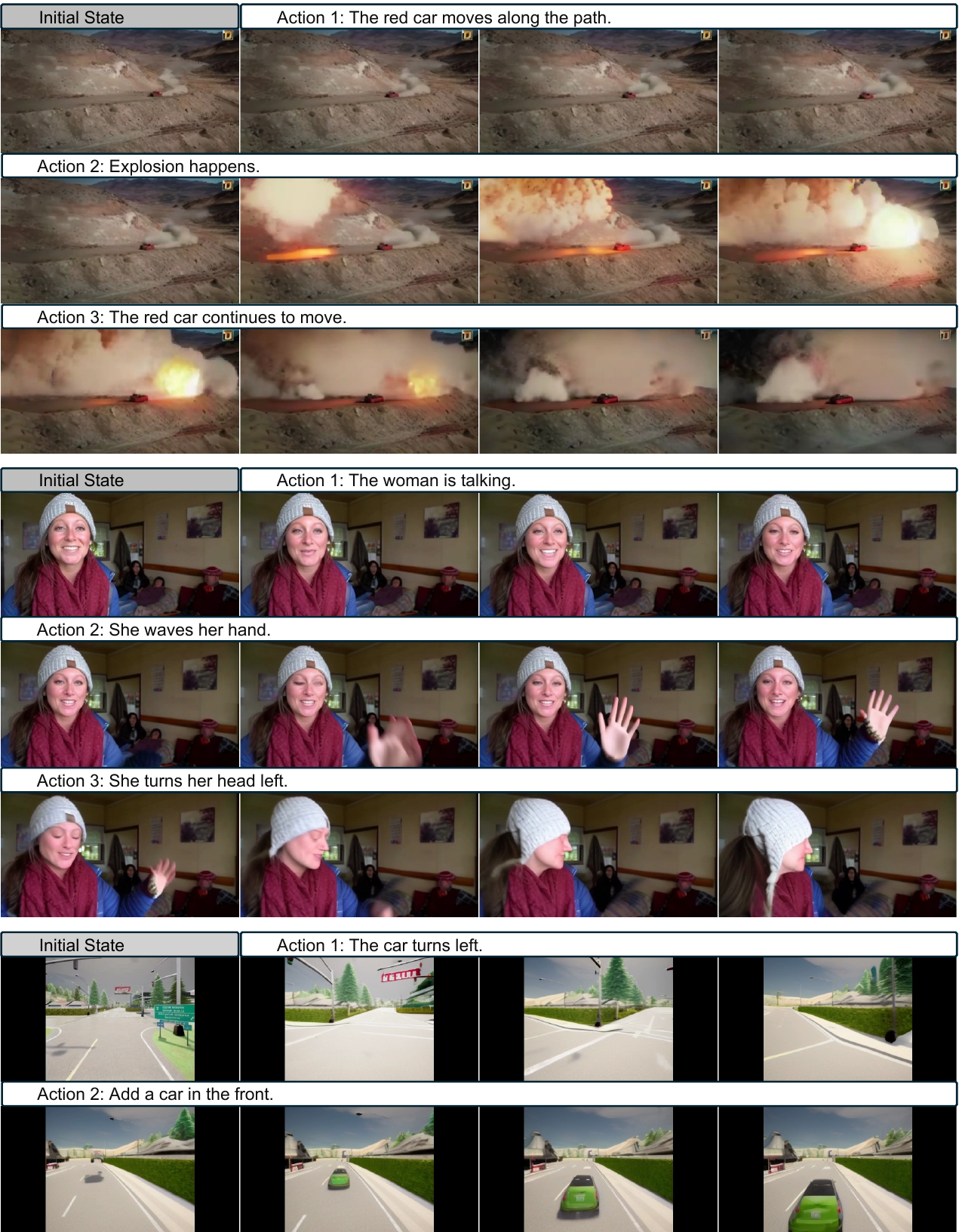}
  \vspace{-45pt}
\end{figure}
\clearpage
\begin{abstract}
  World models simulate future states of the world in response to different actions. They facilitate interactive content creation and provides a foundation for grounded, long-horizon reasoning. Current foundation models do not fully meet the capabilities of general world models: large language models (LLMs) are constrained by their reliance on language modality and their limited understanding of the physical world, while video models lack interactive action control over the world simulations. This paper makes a step towards building a general world model by introducing \modelname, a hybrid autoregressive-diffusion model that simulates world states by generating videos and allows real-time control with free-text actions. \modelname achieves domain {\it generality}, video {\it consistency}, and  {\it controllability} through large-scale pretraining and instruction tuning. Crucially, \modelname bypasses the cost of training-from-scratch by integrating a pretrained LLM (7B) and a pretrained video model, requiring only additional lightweight finetuning. We illustrate extensive outputs by \modelname across diverse domains (indoor/outdoor, natural/urban, human/robot, 2D/3D, etc.). The results indicate great potential of building stronger general world models with larger-scale training.

\end{abstract}

\section{Introduction}

A world model (WM) is an abstract representation that an intelligent system uses to understand and simulate the real world. The model encompasses various aspects of the environment, including physical laws, spatiotemporal knowledge, objects, scenes, agents, and their dynamic interactions. In particular, it allows to predict the future states of the world in response to different actions. Building a general world model, therefore, can serve for {\it interactive content creation}, such as generating realistic virtual scenes for video games and movies, developing immersive experiences in virtual and augmented reality, and creating dynamic simulations for training and educational purposes. Perhaps of even more significance is that a general WM provides a foundation for {\it robust, grounded reasoning} in AI systems, enabling them to anticipate complex environments and plan actions, such as robots navigating disaster scenes safely. WMs also hold the potential to power {\it long-horizon reasoning} that improves decision making in fields like logistics and healthcare, by simulating various scenarios and outcomes and identifying the most effective solutions.

Current {\it large language models (LLMs)} \cite{achiam2023gpt, almazrouei2023falcon, jiang2024mixtral, openai2023gpt4, team2023gemini, touvron2023llama1, touvron2023llama2} are adept at generating human language and are used as surrogates for world models in certain reasoning tasks \citep{hao2023reasoning,wong2023word}. However, language alone is a fundamentally insufficient and inefficient modality for describing various aspects of the world, such as intuitive physics (e.g., predicting fluid flow based on its viscosity) \cite{hu2023language}. Moreover, LLMs lack a robust understanding of physical and temporal dynamics in the real world, relying on patterns in textual data without comprehending the underlying realities they describe \citep{xiang2023language,lai2024vision,mitchell2023debate}. On the other hand, contemporary {\it video generation models} can produce high-quality video content from given initial frames or text prompts \cite{ bao2024vidu, blattmann2023stable, videoworldsimulators2024, esser2023structure,  xing2023dynamicrafter, zhang2024moonshot}. While these models can animate consistent sequences to visualize diverse scenes, they miss the complex interactive nature of the real world, lacking the ability for causal control and intervention with arbitrary actions during simulations. Recent work has also developed interactive world models at scale, such as GAIA-1 \cite{hu2023gaia} for auto-driving, UniSim \cite{yang2023learning} for robotic manipulation, and Genie \cite{bruce2024genie} for 2D games
. These models are typically specific to certain domains, permitting limited sets of actions and/or states.

\begin{figure}[t]
  \centering
  \vspace{-25pt}
  \includegraphics[width=\textwidth]{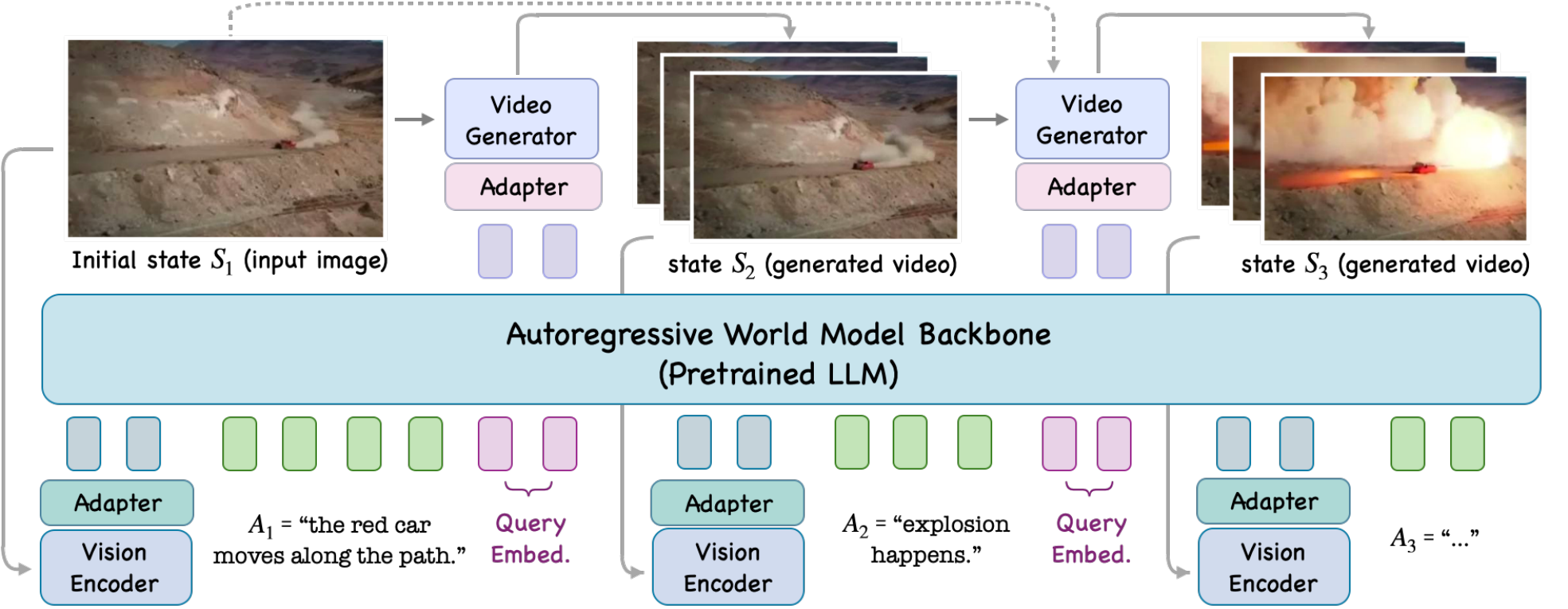}
  \caption{Model architecture of \modelname.}
  \label{fig:arch}
  \vspace{-10pt}
\end{figure}

This work presents \modelname, a step towards a general world model that simulates world states across various domains by generating videos and allows real-time control through arbitrary actions expressed in natural language. 
\modelname is an autoregressive model that sequentially processes actions (free text) and previous states (videos) as inputs and generates next states (videos) as outputs (Figure~\ref{fig:arch}). 
\modelname introduces a staged training strategy akin to the successful recipe of training LLMs \citep{openai2023gpt4,liu2023llm360,wei2021finetuned}
, including: {\it (1) large-scale pretraining} with massive video and text data, respectively, to learn {\it domain-general} understanding of the world and production of {\it consistent} video simulations; and {\it (2) instruction tuning} with high-quality text-video sequential data to learn any-time text {\it controllability} during video generation.

Crucially, the pretraining stage allows for the separate training of text and video modeling. We thus can simply reuse existing pretrained LLMs and (text-to-)video generation models that have already achieved domain generality and video consistency in their own pretraining. We only need to stitch and align the language and video models together with necessary additional modules and lightweight tuning as described in \S\ref{sec:method}. More specifically, in this work, we use the Vicuna-7B-v1.5 language model~\cite{vicuna2023} and the DynamiCrafter text-to-video model~\cite{xing2023dynamicrafter} as the backbone. Using larger, more sophisticated pretrained models (such as GPT-4 and Sora) is expected to yield stronger performance. For the instruction tuning stage, we craft a large diverse set of action-state sequential data, by re-captioning general-domain videos and synthesizing with various simulators for robots, in-/out-door activities, driving, 2D games, and more. Similar to instruction tuning of LLMs that boosts their instructability in general unseen domains, tuning on the curated data boosts the world model's real-time controllability that generalizes to broad unseen states and actions.

We illustrate extensive outputs generated by \modelname across various domains in \S\ref{sec:exp}. The model demonstrates a range of desirable properties not exhibited by previous models. The results also indicate great potential for further enhancement with larger-scale training in the future.

\begin{addmargin}[-15pt]{0pt} 
\begin{itemize}\setlength\itemsep{0pt}

\item {\bf The model simulates video states across broad domains: }
\modelname is capable of generating videos across a wide range of general domains, such as indoor/outdoor, natural/urban, human/robot, 2D/3D, and other scenarios. This domain generality is primarily due to the large-scale video pretraining (inherited from the pretrained video model).

\item {\bf The model permits on-the-fly control with free-text actions: }
\modelname accepts natural language actions as inputs during video generation to direct future world states. This differs crucially from previous text-to-video models which allow text prompts only at the beginning of the video. The on-the-fly control fulfills the promise of the world model to support interactive content generation and enhance robust reasoning and planning. The capability is enabled by the autoregressive architecture of the model (which permits text inputs at any time), the pretrained LLM backbone (which understands any text expressions), and the instruction tuning stage (which substantially enhances the effectiveness of control).

\item {\bf Action controllability transfers across domains: }
As above, instruction tuning with high-quality data allows the model to learn effective action control and transfer to different unseen domains. We demonstrate that actions learned from a specific domain apply seamlessly to states in diverse new domains.

\item {\bf Autoregressive model backbone enables longer videos: }
Existing video generation models based on diffusion architectures typically produce videos of a fixed length (e.g., 2 seconds). By integrating the pretrained video model with the LLM autoregressive backbone, \modelname is capable of extending the video duration indefinitely in an autoregressive manner. Together with the additional training (e.g., instruction tuning), we show \modelname can generate longer videos (e.g., 8 seconds) of higher quality.




\end{itemize}
\end{addmargin}





 









\section{Methods}\label{sec:method}

\subsection{Model Architecture}
\modelname is an autoregressive world model.
Given the previous states of the world, e.g., images or video clips, and a natural language action, it predicts the next state of the world, which is also a video clip.
Specifically, it formulates a state transition distribution: 
\begin{equation}
\begin{split}
    P(S_{t} \mid S_{t-1}, A_{t-1}, ..., S_1, A_1),
\end{split}
\end{equation}
where $S_i$ and $A_i$ are the state and action at time step $i$, respectively. Each state $S_i = (s_1, ..., s_N)$ is a single or a sequence of video frames, and each action $A_i = (x_1, ..., x_M)$ is a sequence of text tokens. At the first time step, the state $S_1$ is one single image, and the states $S_2, ..., S_N$ at the following steps are video clips. 

Figure~\ref{fig:arch} gives an overview of the model architecture. The two core components of of \modelname include the autoregressive backbone, which stems from a pretrained LLM, and the video generator, which is initialized with a pretrained video model. To stitch the two components together, other necessary components are added, including a vision encoder, and two adapters connecting the vision encoder to the LLM backbone, and the LLM backbone to the video generator, respectively.

At each time step $t$, the autoregressive backbone accepts three sets of embedding vectors as inputs: (1) the first is the sequence of visual embeddings, by the vision encoder followed by the adapter (a Q-Former~\cite{li2023blip}), that encodes the previous world state $S_{t-1}$; (2) the second is the token embeddings of the text words in action $A_{t-1}$; and (3) the third is a sequence of learnable embedding vectors (a.k.a. query embeddings). The length and positions of the query embeddings correspond exactly to those of the output embeddings by the autoregressive backbone to be fed to the video generator. Intuitively, the query embeddings stimulate the model to start generating videos \cite{dong2023dreamllm}. The autoregressive backbone then generates a sequence of output embeddings. 
The adapter, which is a Q-Former, accepts the output embeddings and produces a new sequence of embeddings. Finally, the video generator takes the embeddings and generates the video clip outputs $S_t$. To improve the consistency of the new video clip with the preceding video clip $S_{t-1}$, the video generator additionally takes the last four frames of $S_{t-1}$ as input (or the single image of $S_{t-1}$ as input if $S_{t-1}$ is the initial state $S_1$). In addition, the video generator will take an FPS number to control the motion level of the video. The number of frames generated in each video clip $S_t$ depends on the specific pretrained video model used for initializing the video generator. As described below, we used the DynamiCrafter~\cite{xing2023dynamicrafter} which generates 16 frames.

\subsection{Staged Training}

A general world model needs to achieve {\it consistency}, {\it controllability}, and {\it generality}---it needs to generate consistent videos to describe the world state accurately, allow on-the-fly control by accepting natural language actions at any time during video generation, and perform the above well across all diverse domains (with different scenes and actions). 

To this end, direct training of the world model requires massive high-quality (video $S_1$, text $A_1$, video $S_2$, $\dots$) sequences as training data, which is hard to obtain in practice. We instead devise a two-stage training strategy consisting of {\it pretraining} and {\it instruction tuning}. 

The pretraining stage aims to acquire a few key capabilities, including (1) consistent general video generation of the video generator, (2) general text understanding of the autoregressive backbone to process actions, and (3) alignment of the  representation spaces between the two components. The first two capabilities can be learned separately by training the video generator and the autoregressive backbone individually, or even by just plugging in existing pretrained video models and LLMs that already possess these capabilities during their own pretraining. The reuse of separately pretrained video and language models significantly reduces the training costs of the world model.


In the instruction tuning stage, we train the model on a curated video dataset with high-quality instructions (actions) that focus on the dynamics of the videos. This training is aimed at enhancing the model's ability to follow natural language instructions and accurately predict subsequent video states based on these directions.

We describe more details of the two training stages in the next sections, respectively.

\subsubsection{Pretraining for Domain Generality and Generation Consistency}

The pretraining stage aims to achieve the core capabilities of {\it consistency} and {\it generality} as described above. This is similar to the process of building an LLM where large-scale pretraining enables the LLM to generate consistent/fluent text in general domains. 

General understanding of natural language can be achieved by massive training on text data, and generation of consistent general videos can be achieved by massive training on video data. Both of these have been done on existing pretrained LLMs and video generation models. We thus can directly reuse these models.

Specifically, we use Chat-Univi~\cite{jin2023chat}, which is a Vicuna-7B-v1.5 LLM equipped with a vision encoder, as our base LLM (autoregressive backbone), and DynamiCrafter~\cite{xing2023dynamicrafter} as our base video generation model. DynamiCrafter is a diffusion model pretrained to generate a video given an image and a text prompt.

In the pretraining stage, we additionally want to align the pretrained LLM backbone and video model, so that the output embeddings from the LLM can be passed along to the video model as input for video generation. We used a video caption dataset WebVid-10M~\cite{bain2021frozen} for the alignment training. For each (video, caption), we feed the first frame of the video and the caption into the LLM+vision encoder and get the output embeddings from the LLM. Meanwhile, we feed the caption into the text encoder of the video generation model and get the caption embeddings. We aim to match the two embeddings, so that the output embeddings from LLM can be understood by the vidoe  generation model (just as how it understands the embeddings from its text encoder). Specifically, we minimize the L2 loss between the two sets of embeddings, and trains the parameters of the adapter between the LLM and video generator, as well as the query embeddings. Both the pretrained LLM and video generation model are fixed at this stage.

\subsubsection{Instruction Tuning for Real-Time Controllability}\label{sec:instruction_tuning_data}

This stage aims to gain real-time controllability by training the model on high-quality instruction tuning data. We construct such a dataset, which contains captions to precisely describe the dynamics of different clips in each video. With the data, we finetune the model by minimizing a diffusion loss on the videos given the instructions. In this stage, both the video generator and query embeddings are finetuned, while other components are fixed. 

Below we describe the creation of the instruction tuning data in more details. An overview of the collected data is summarized in Table~\ref{tab:data}. The data come from both public corpus and simulators with careful data processing.


\paragraph{Public Video Datasets} 
To make the dataset general, we use a large-scale video dataset, Panda-70M \cite{chen2024panda70m}. 
We first filter the dataset by aesthetic score evaluation, optical flow magnitude assessment, cut detection, static video detection, and clip length filtering. 
Different from previous text-to-video models, our model emphasizes the controllability of natural language actions towards the next state. Therefore, we do re-captioning of the videos to get better captions that focus on the dynamics of the videos. we prompt GPT-4 Turbo \cite{achiam2023gpt} to generate captions describing the dynamics of four frames sampled from each video clip. This process yields a total of 500k video-text pairs. Besides Panda-70M, we also collect video-action pairs from existing action-annotated datasets, including Something-Something V2~\cite{something-something-v2}, BridgeData V2~\cite{bridge_data}, and EPIC-KITCHENS~\cite{epic-kitchen}. This includes 260k examples.

\begin{wraptable}{r}{0.5\textwidth}
  \centering
  \small
  \begin{tabular}{c|c|c}
    \toprule
    Dataset & Category & \#Videos \\
    \midrule
    Panda-70M & YouTube & 500k \\
    Something-Something & Human Activity & 188k \\
    BridgeData V2 & Robot Arm & 33k \\
    HM3D & Indoor & 152k \\
    MP3D & Indoor & 70k \\
    StreetLearn & Street view & 146k \\
    Carla & Driving & 75k \\
    Coinrun & 2D Game & 30k \\
    EPIC-KITCHENS & Kitchen & 39k \\
    \midrule
    Total & - & 1.2M \\
    \bottomrule
  \end{tabular}
  \caption{Instruction Tuning data statistics.}
  \label{tab:data}
\end{wraptable}

\paragraph{Simulation Data} 
To provide our model with more diverse and accurate training experience, we use simulation environments to collect video-action pairs. 
CARLA~\cite{carla} is a simulation platform for autonomous driving. It supports flexible modifications to the environment at runtime, making it suitable for simulating unexpected actions, such as \textit{Change the weather to Sunset} or \textit{Add a car to the front}. We sampled 75k video-action pairs from Carla. MP3D~\cite{Matterport3D} and StreetLearn~\cite{street-learn, street-learn2} are indoor and urban panorama scans. We built simulation environments to render these 3D scans. Turning actions such as \textit{turn right for 60 degrees} can be constructed by gradually changing camera poses and collecting corresponding image projections. Besides, we prompt GPT-4 Turbo to generate scene descriptions, so that the instructions include both turning actions and the final scene descriptions. We got 70k data from MP3D and 146k data from StreetLearn, respectively. HM3D~\cite{hm3d} is a 3D environment dataset of real-world indoor scenes. We used Habitat-Lab~\cite{habitat1,habitat2,habitat3} to render these indoor scenes and collect data by sampling trajectories randomly. We created 152k data from it. Finally, we used Coinrun~\cite{coinrun} for collecting 2D game simulation data, resulting in 30k data.

\section{Qualitative Results}\label{sec:exp}
We show qualitative results that demonstrate the core capabilities of \modelname as a world simulator. Readers are encouraged to refer to \url{https://world-model.ai} for live video examples. We aim to report more quantitative results in the future.

\subsection{On-The-Fly Control across Domains}
\begin{figure}[!ht]
    \centering
    \begin{subfigure}[b]{1\textwidth}
        \centering
        \includegraphics[width=\textwidth]{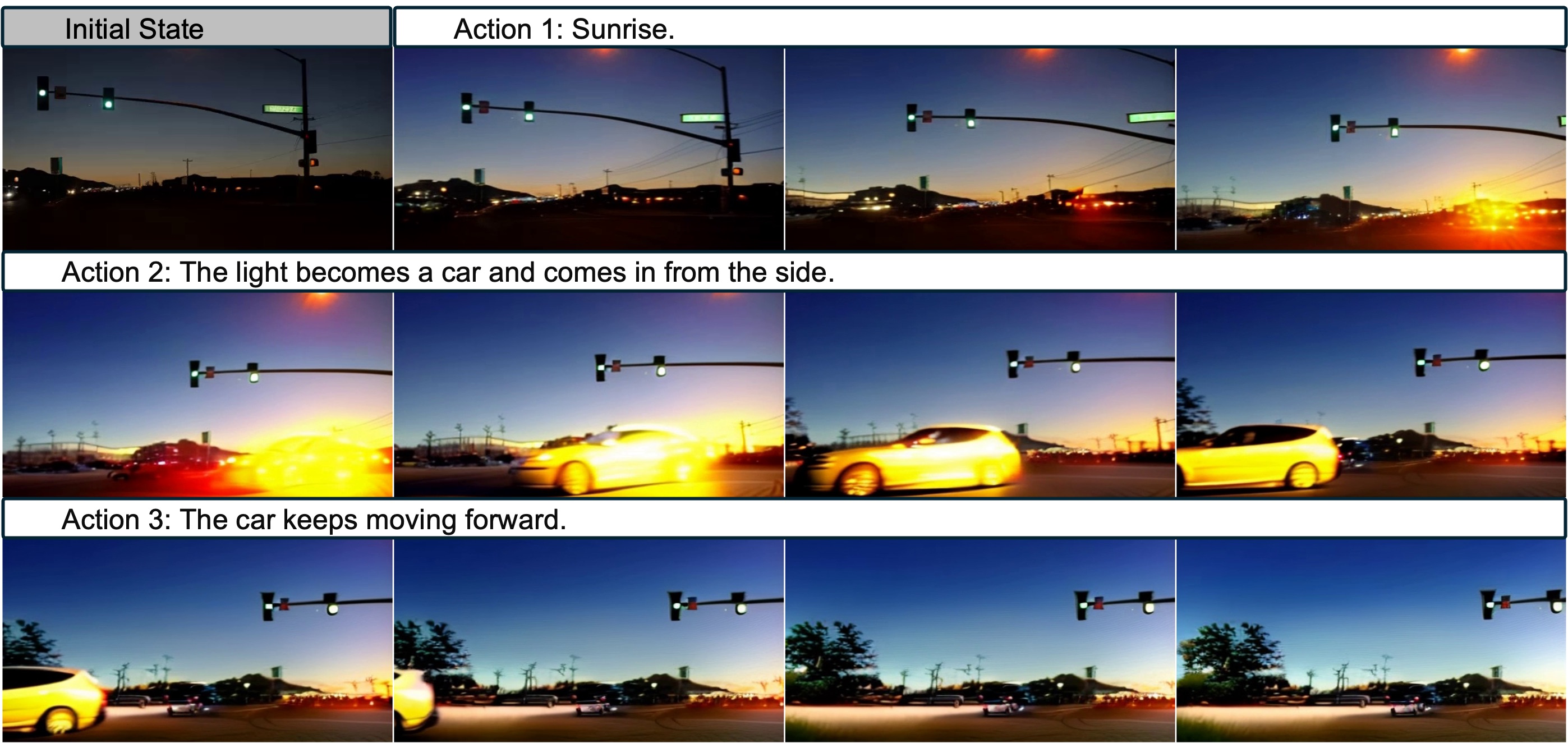}
        \caption{Sci-fi style movie scene.}
    \end{subfigure}
    
    \begin{subfigure}[b]{1\textwidth}
        \centering
        \includegraphics[width=\textwidth]{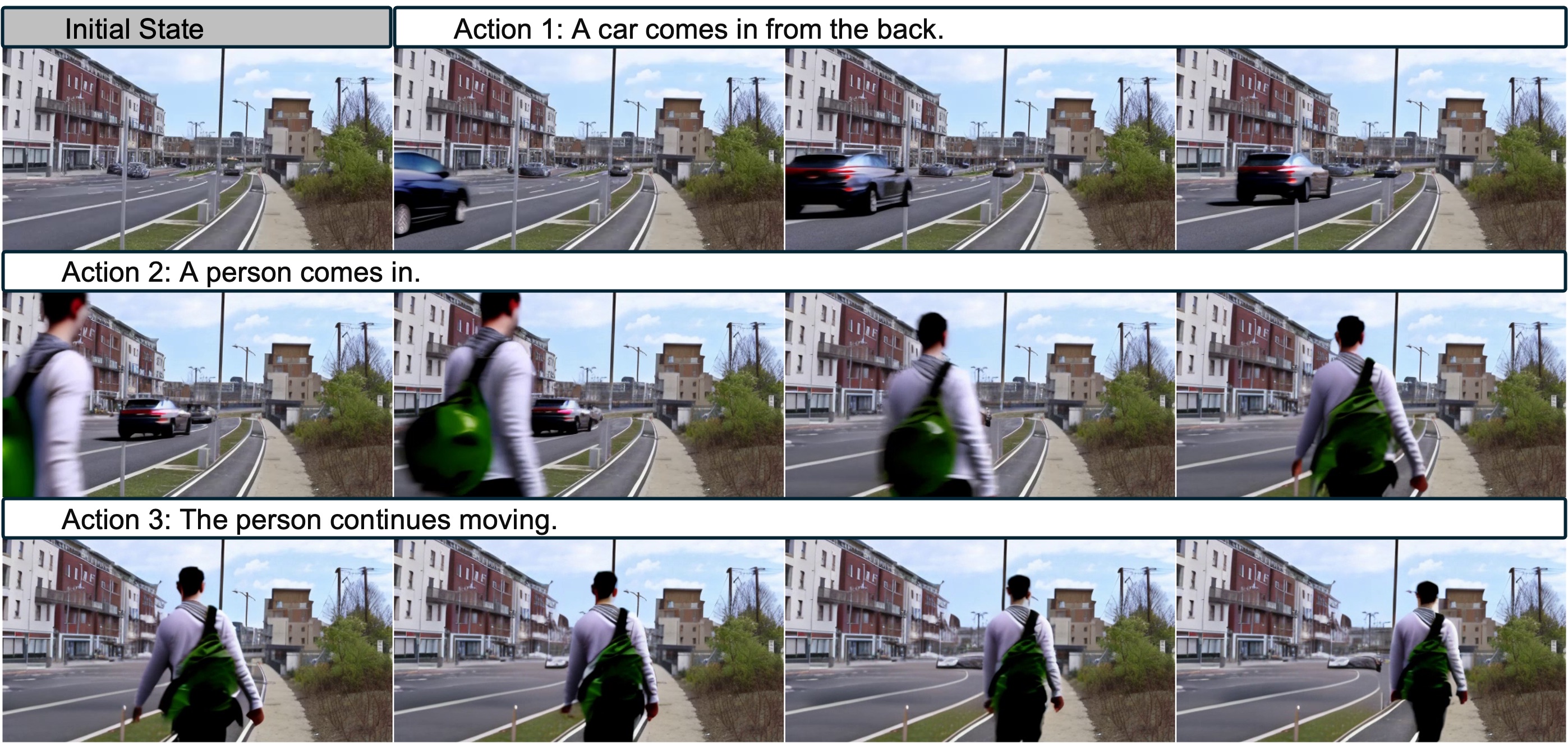}
        \caption{Everyday city scene.}
    \end{subfigure}
    \begin{subfigure}[b]{1\textwidth}
        \centering
        \includegraphics[width=\textwidth]{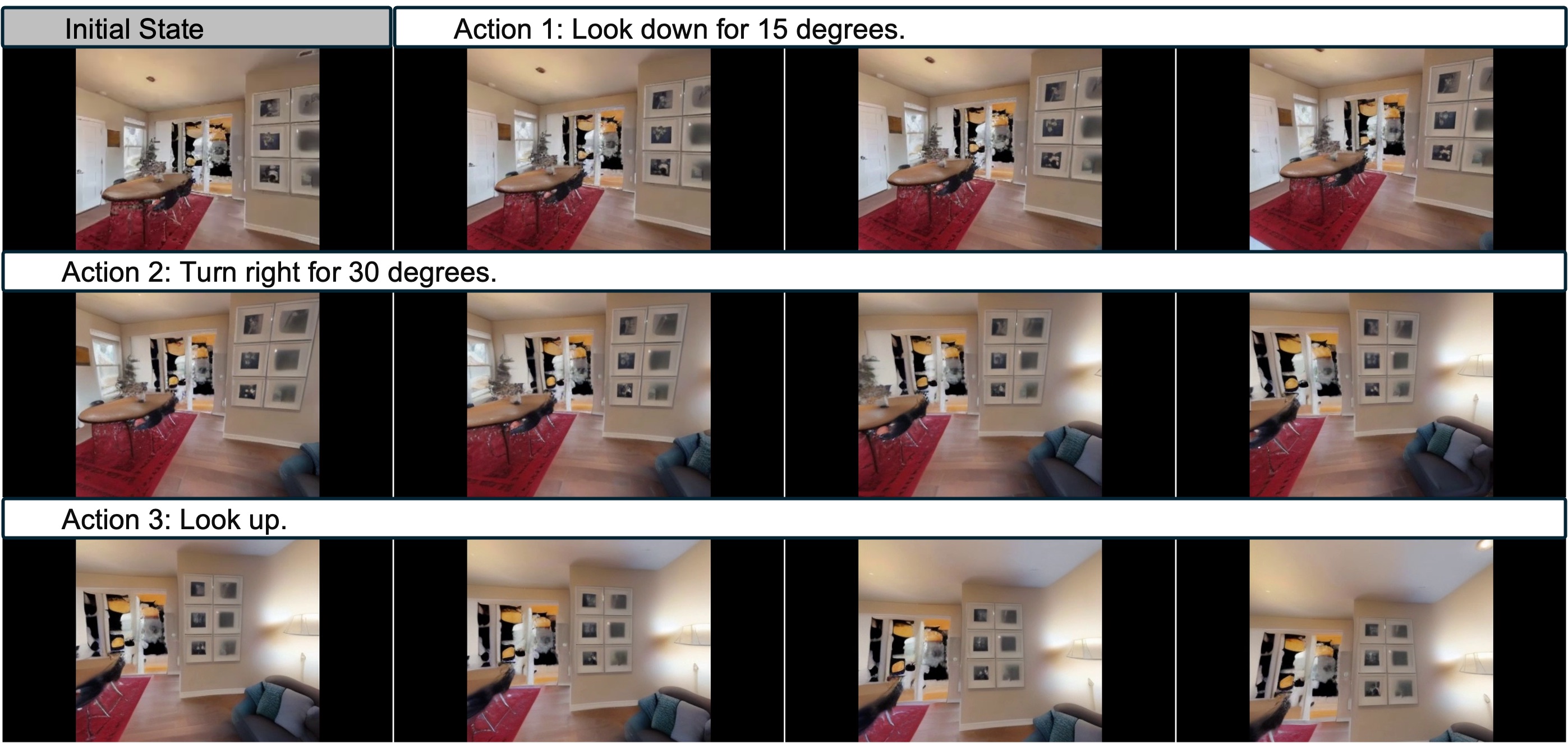}
        \caption{Household scene.}
    \end{subfigure}
    \caption{\modelname can generate sci-fi or real-life videos in both indoor and outdoor environments.}
    \label{fig:2d3d}
\end{figure}

\begin{figure}[!ht]
    \centering
    \begin{subfigure}[b]{1\textwidth}
        \centering
        \includegraphics[width=\textwidth]{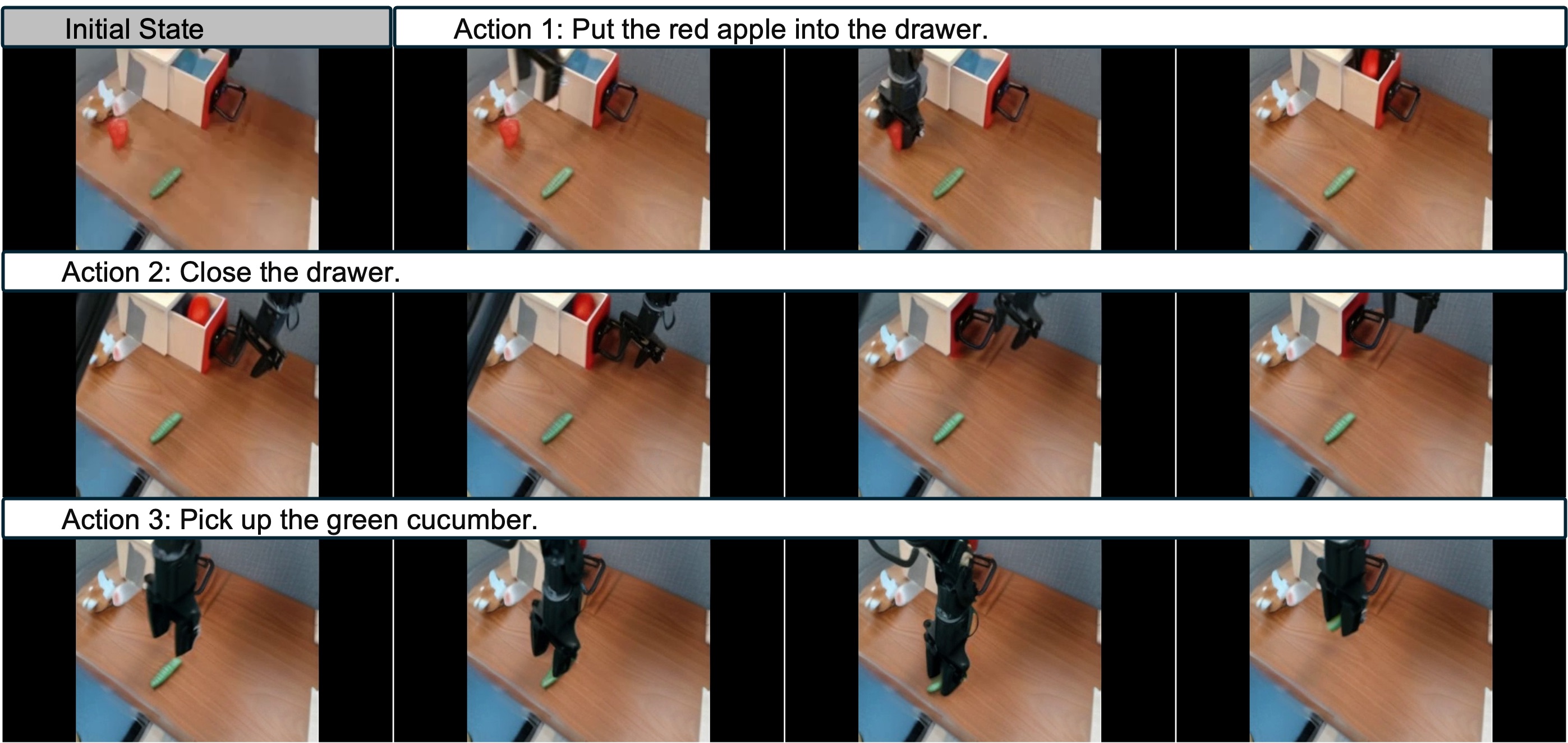}
        \caption{A robotic arm manipulating several stuff.}
    \end{subfigure}
    
    \begin{subfigure}[b]{1\textwidth}
        \centering
        \includegraphics[width=\textwidth]{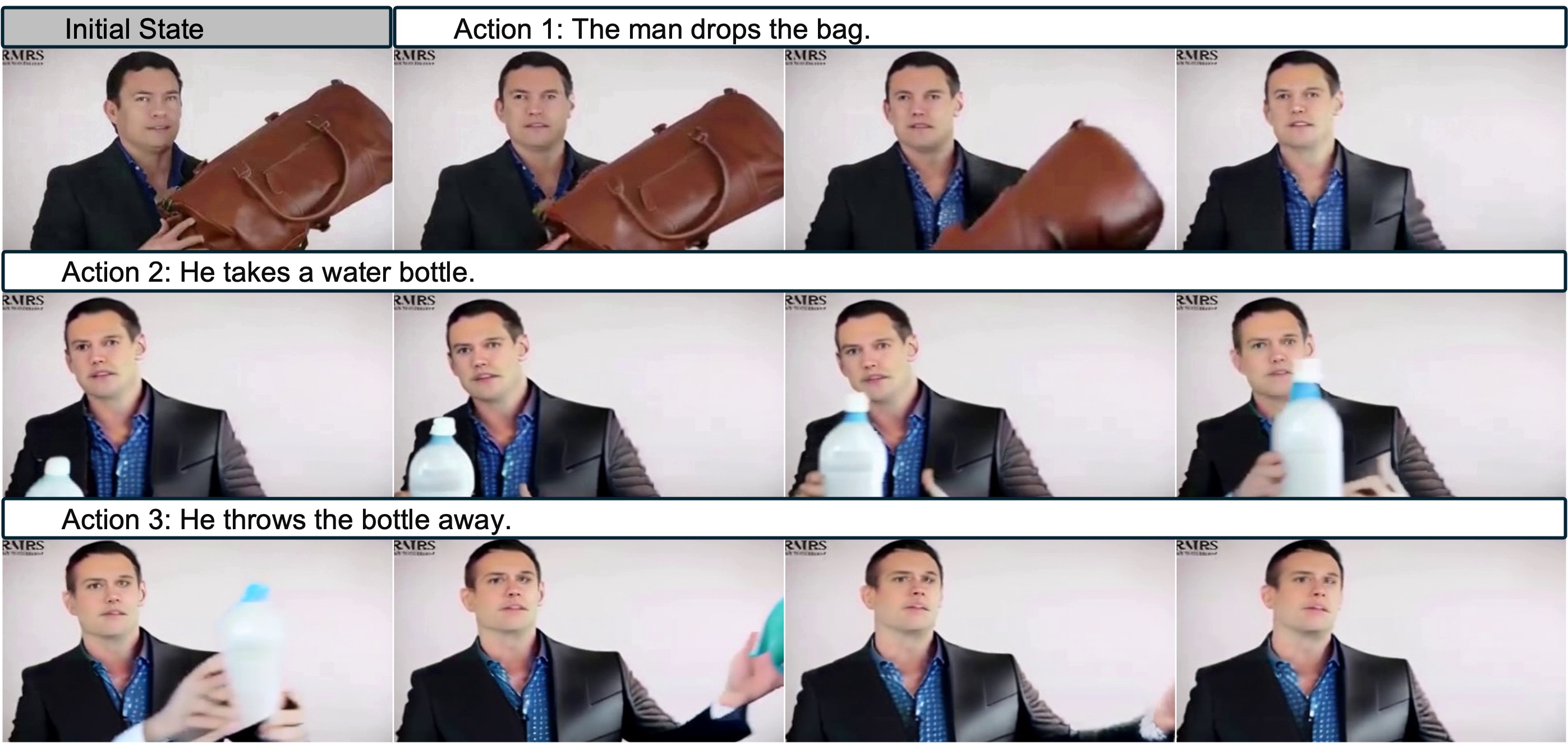}
        \caption{A real human doing some actions.}
    \end{subfigure}
    \begin{subfigure}[b]{1\textwidth}
        \centering
        \includegraphics[width=\textwidth]{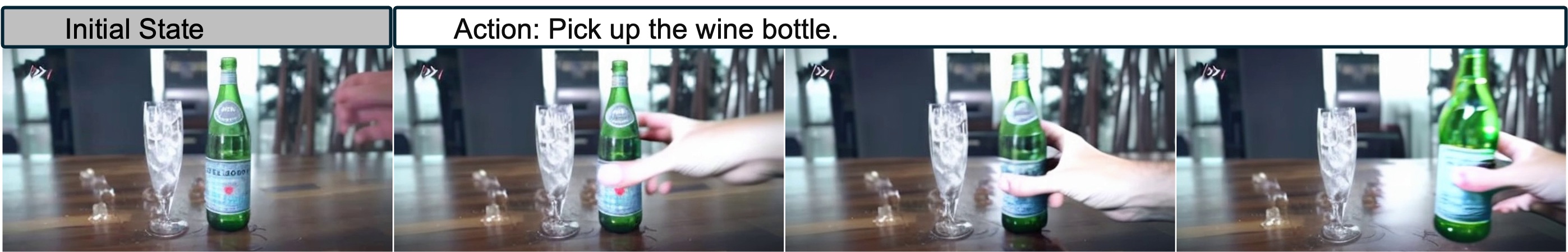}
        \caption{An ego-view human activity}
    \end{subfigure}
    \caption{\modelname is capable of generating both robotics and human videos.}
    \label{fig:human_robot}
\end{figure}

\begin{figure}[!ht]
    \centering
    \begin{subfigure}[b]{1\textwidth}
        \centering
        \includegraphics[width=\textwidth]{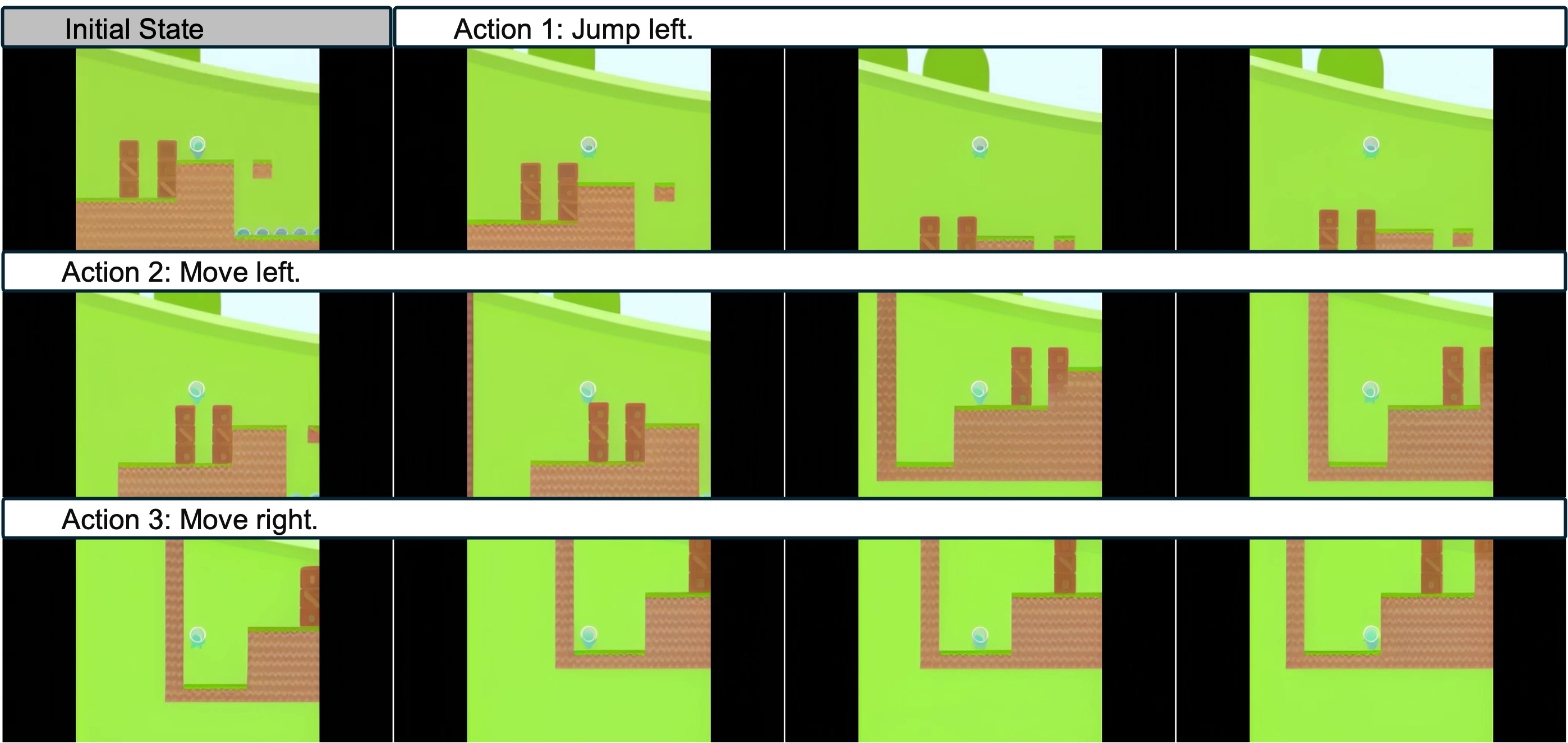}
        \caption{2D game Coinrun.}
    \end{subfigure}
    
    \begin{subfigure}[b]{1\textwidth}
        \centering
        \includegraphics[width=\textwidth]{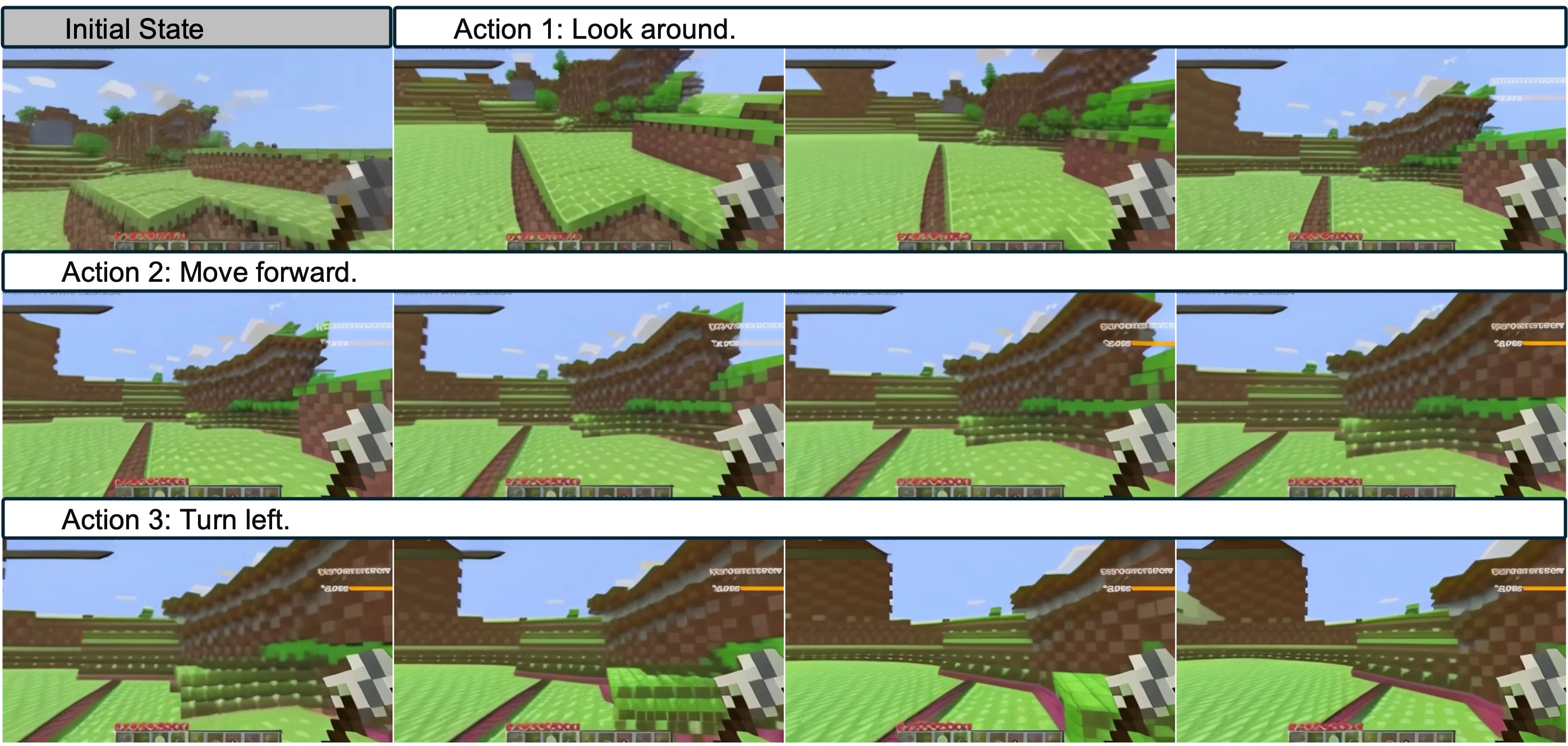}
        \caption{3D game Minecraft.}
    \end{subfigure}
    \begin{subfigure}[b]{1\textwidth}
        \centering
        \includegraphics[width=\textwidth]{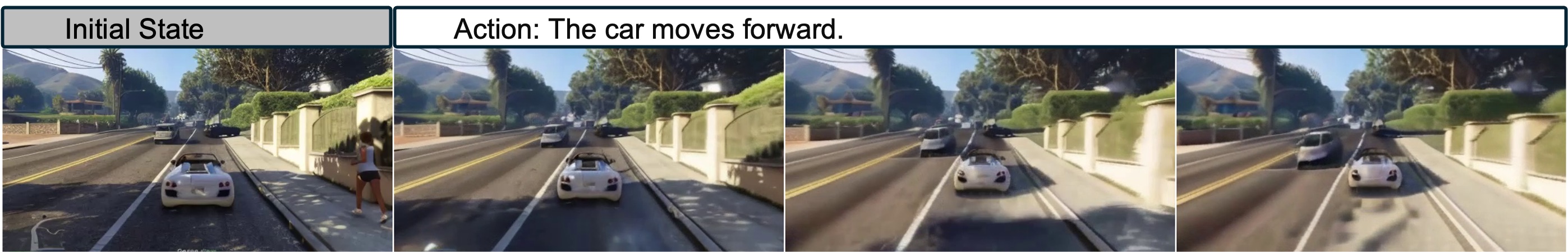}
        \caption{3D game Grand Theft Auto V.}
    \end{subfigure}
    \caption{\modelname can generate various 2D and 3D game videos.}
    \label{fig:game}
\end{figure}

\begin{figure}[!ht]
    \centering
    
    \begin{subfigure}[b]{1\textwidth}
        \centering
        \includegraphics[width=\textwidth]{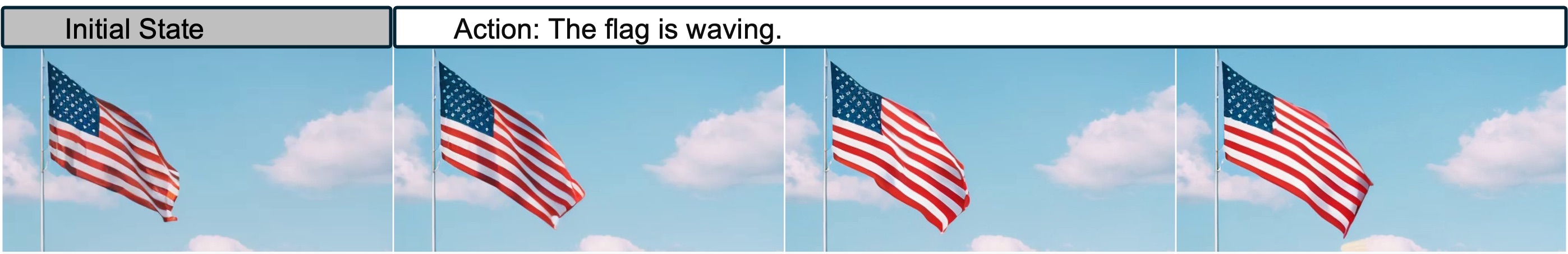}
        \caption{Object movements because of wind.}
    \end{subfigure}

    \begin{subfigure}[b]{1\textwidth}
        \centering
        \includegraphics[width=\textwidth]{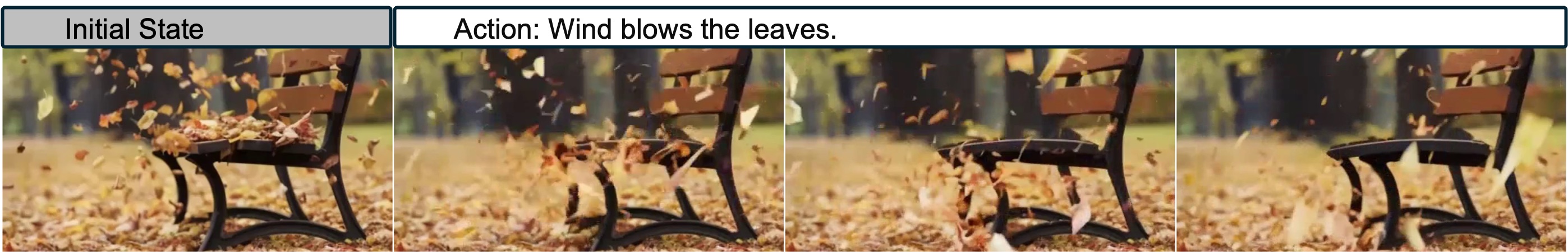}
        \caption{Object movements because of wind.}
    \end{subfigure}
    
    \begin{subfigure}[b]{1\textwidth}
        \centering
        \includegraphics[width=\textwidth]{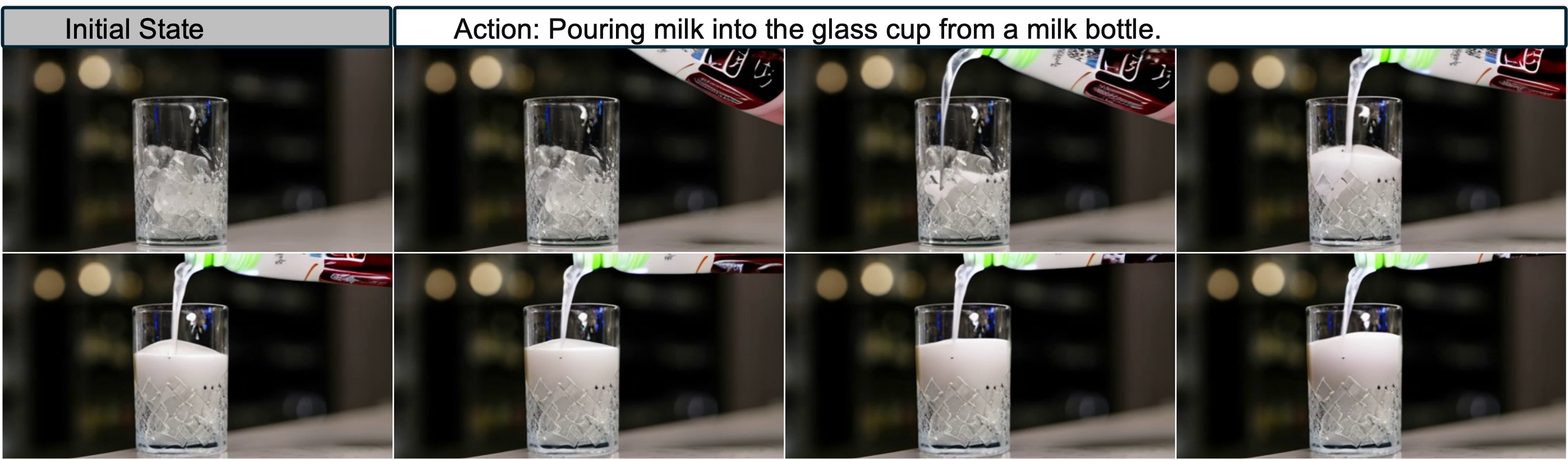}
        \caption{The flow of liquid.}
    \end{subfigure}

    \begin{subfigure}[b]{1\textwidth}
        \centering
        \includegraphics[width=\textwidth]{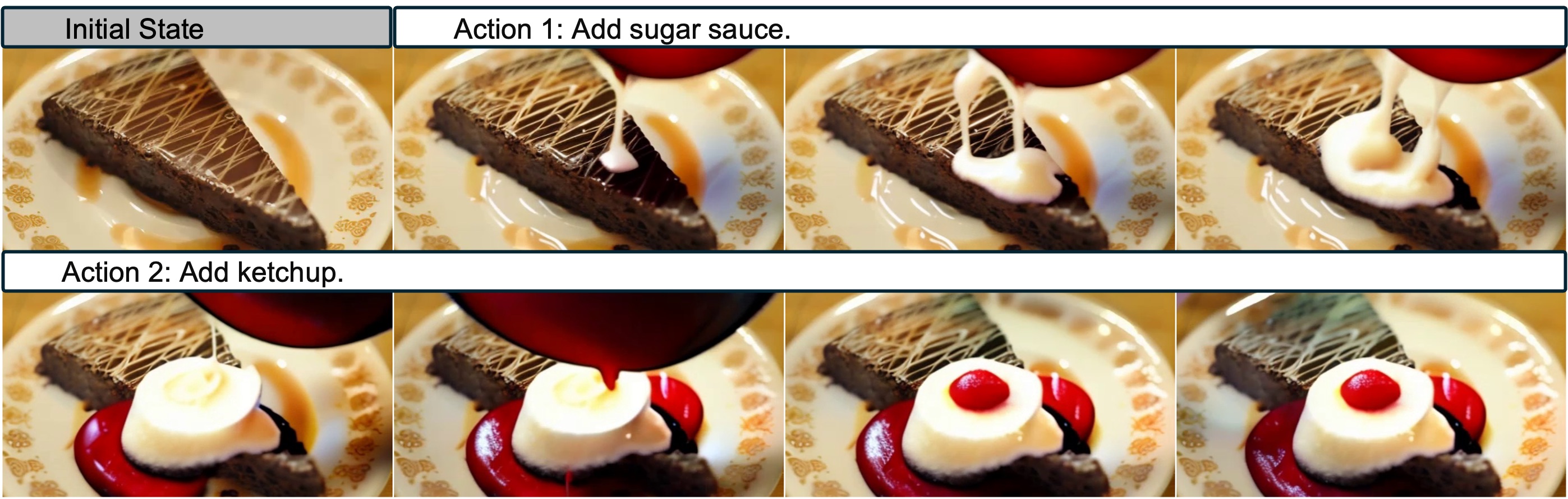}
        \caption{The flow of viscous liquid.}
    \end{subfigure}

    \caption{\modelname is capable of generating videos that include common physical phenomena.}
    \label{fig:physics}
\end{figure}

\modelname is a general world model capable of generating videos across a broad range of domains. It permits on-the-fly control with free-text actions, \emph{i.e.}, it can accept text action control anytime during the video generation and predict future world states accordingly. We show the generation results of indoor/outdoor videos in Figure~\ref{fig:2d3d}, robot/human videos in Figure~\ref{fig:human_robot}, and 2D/3D game videos in Figure~\ref{fig:game}. In Figure~\ref{fig:physics}, we also show videos that correctly demonstrate basic physical phenomena, demonstrating the model's understanding of real-world physical concepts.

\subsection{Action Controllability Transfer}
\begin{figure}[!ht]
    \centering

    \begin{subfigure}[b]{1\textwidth}
        \centering
        \includegraphics[width=\textwidth]{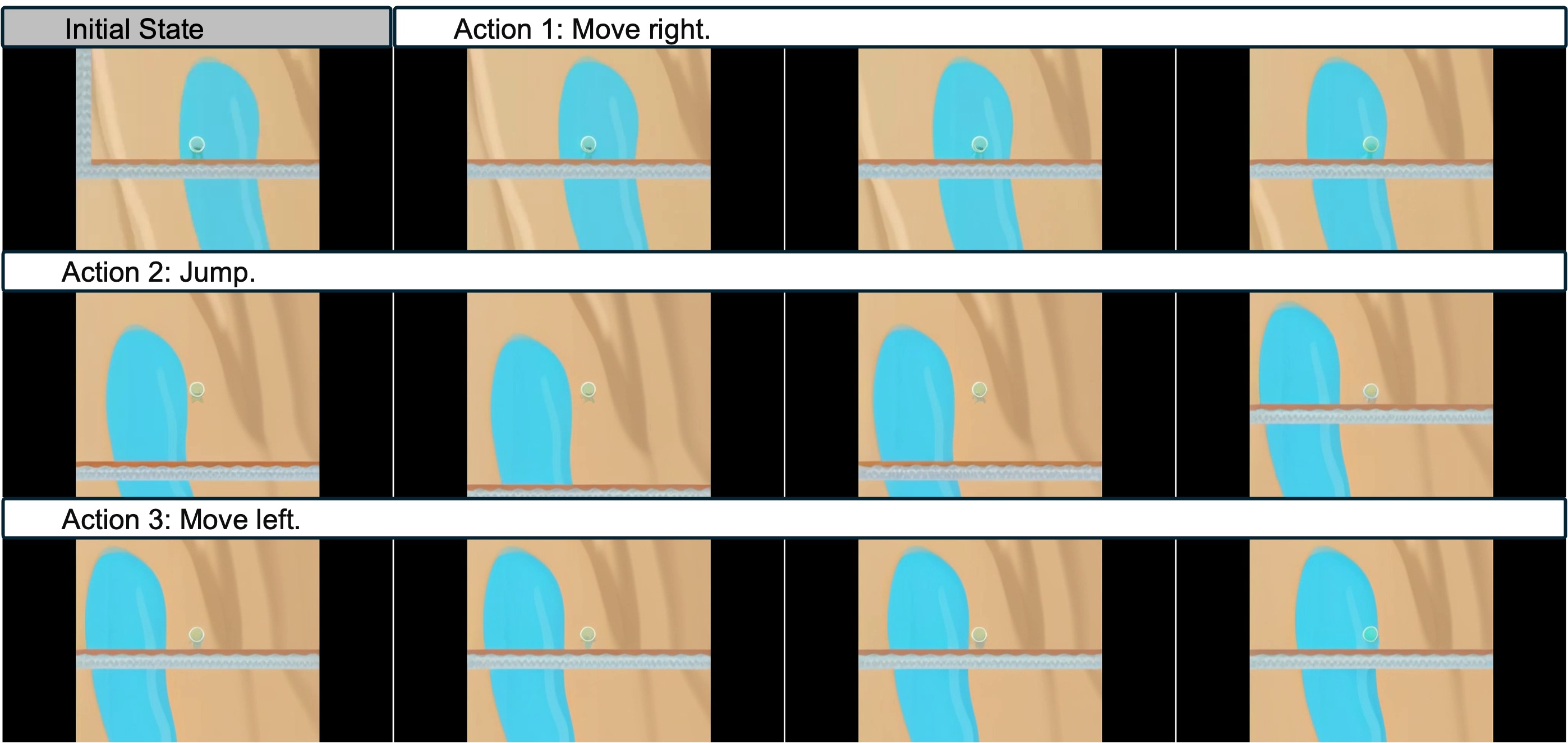}
        \caption{Source domain (Coinrun).}
    \end{subfigure}
    
    \begin{subfigure}[b]{1\textwidth}
        \centering
        \includegraphics[width=\textwidth]{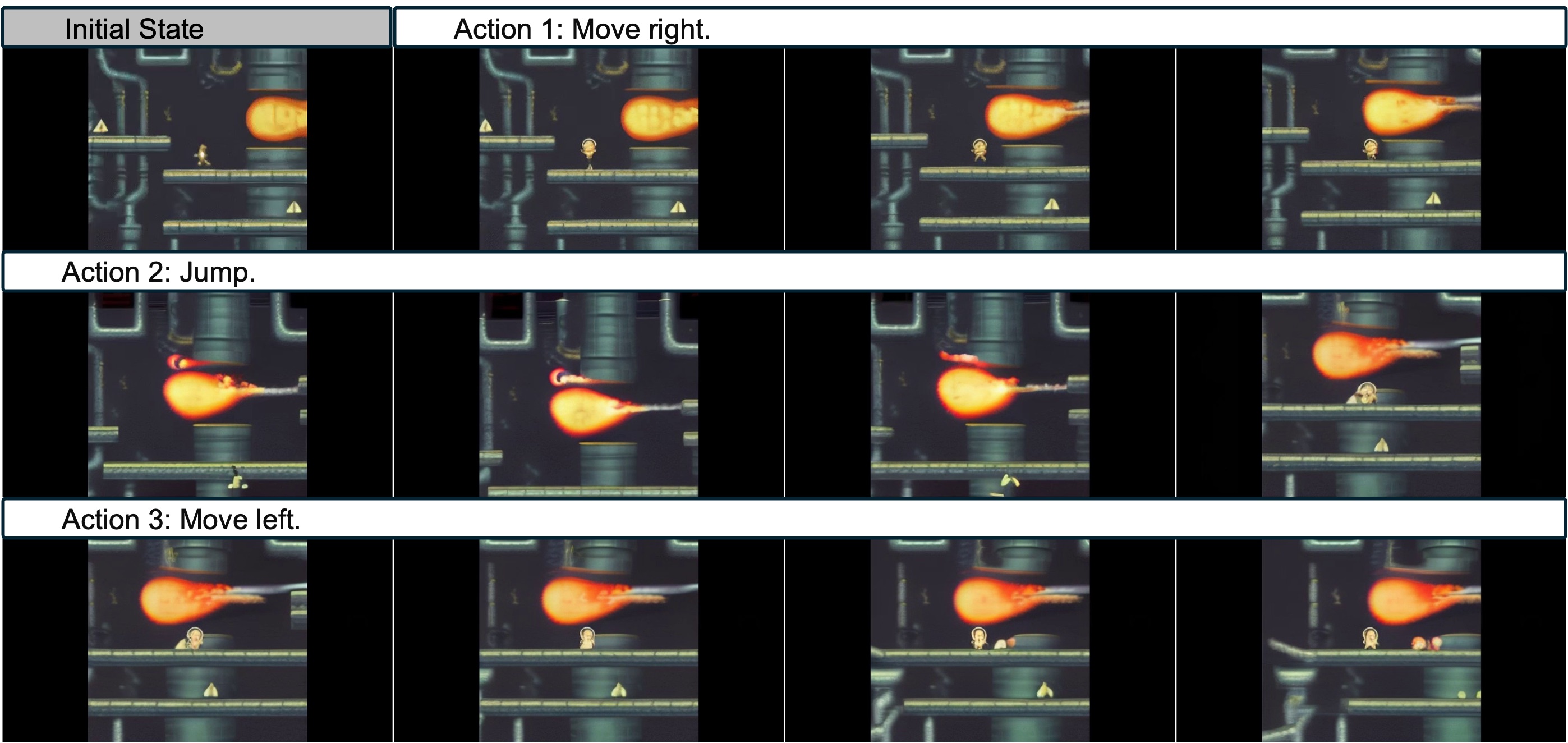}
        \caption{Target domain (Game 1).}
    \end{subfigure}
    
    \begin{subfigure}[b]{1\textwidth}
        \centering
        \includegraphics[width=\textwidth]{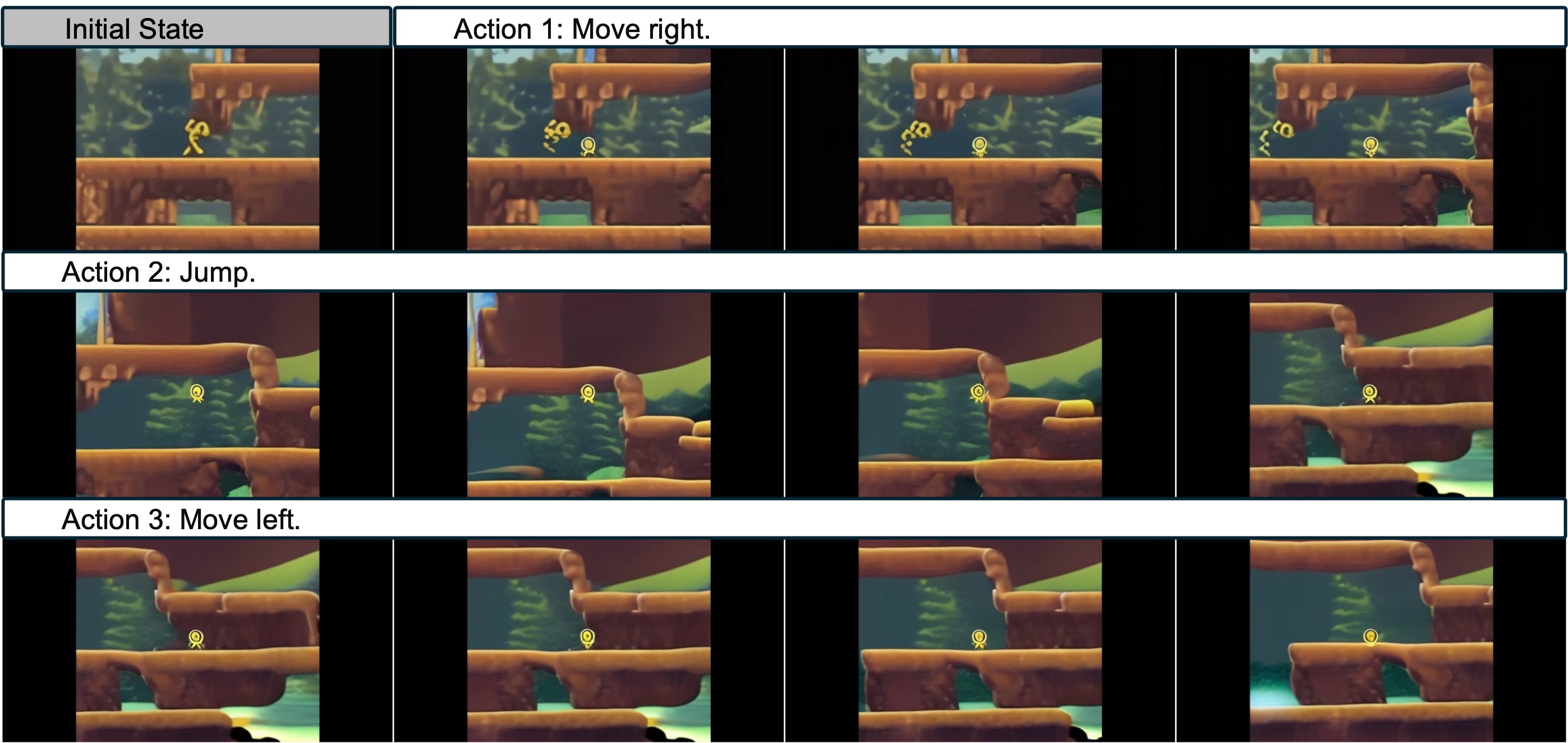}
        \caption{Target domain (Game 2).}
    \end{subfigure}
    \caption{\modelname transfers the 2D game ability from the only 2D game in our training data, Coinrun, to other unseen 2D games.}
    \label{fig:transfer_coinrun}
\end{figure}

\begin{figure}[!ht]
    \centering
    
    \begin{subfigure}[b]{1\textwidth}
        \centering
        \includegraphics[width=\textwidth]{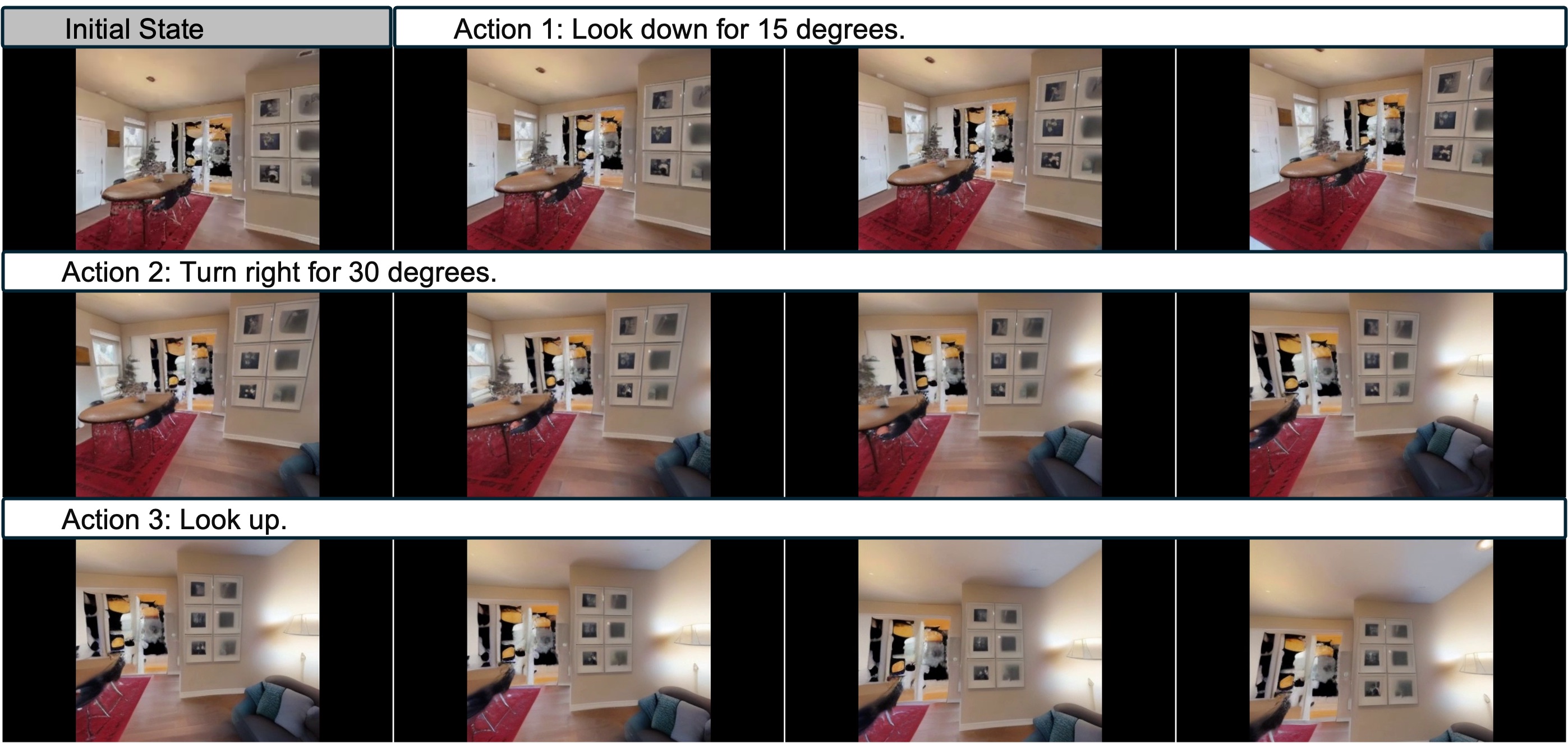}
        \caption{Source domain (HM3D).}
    \end{subfigure}
    \begin{subfigure}[b]{1\textwidth}
        \centering
        \includegraphics[width=\textwidth]{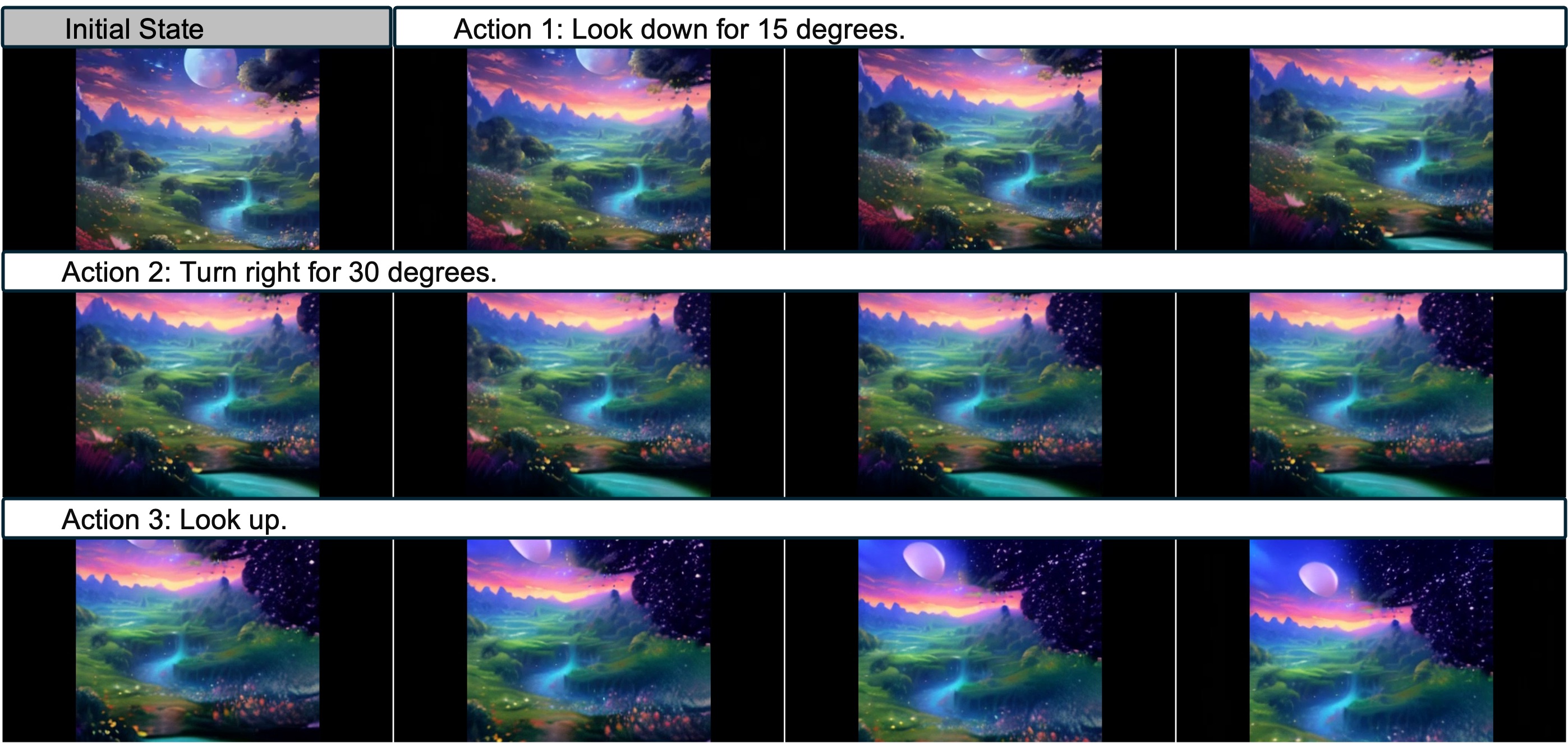}
        \caption{Target domain (Magic scene).}
    \end{subfigure}

    \begin{subfigure}[b]{1\textwidth}
        \centering
        \includegraphics[width=\textwidth]{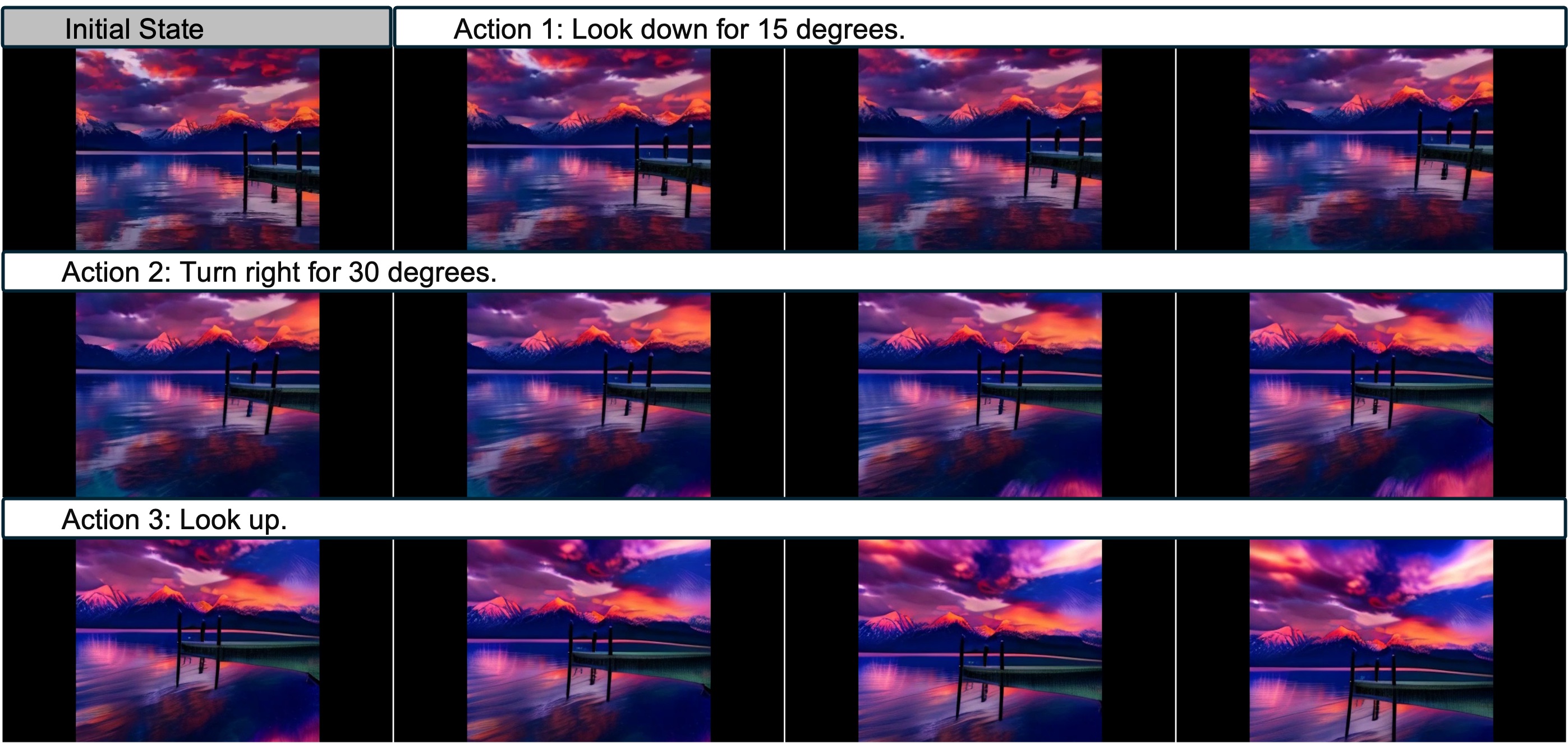}
        \caption{Target domain (Colorful nature scene).}
    \end{subfigure}

    \caption{\modelname transfers the 3D indoor simulator ability to other unseen domains.}
    \label{fig:transfer_hm3d}
\end{figure}

Although some actions and their corresponding motion patterns only appear in some of the simulation data, we found that \modelname can transfer the action controllability to different unseen domains. As shown in Figure~\ref{fig:transfer_coinrun} and Figure~\ref{fig:transfer_hm3d}, \modelname transfers 2D game ability from Coinrun and 3D simulator ability from HM3D to other unseen domains, respectively.

\subsection{Autoregressively Generating Longer Videos} 
\begin{figure}[!ht]
    \centering
    \begin{subfigure}[b]{1\textwidth}
        \centering
        \includegraphics[width=\textwidth]{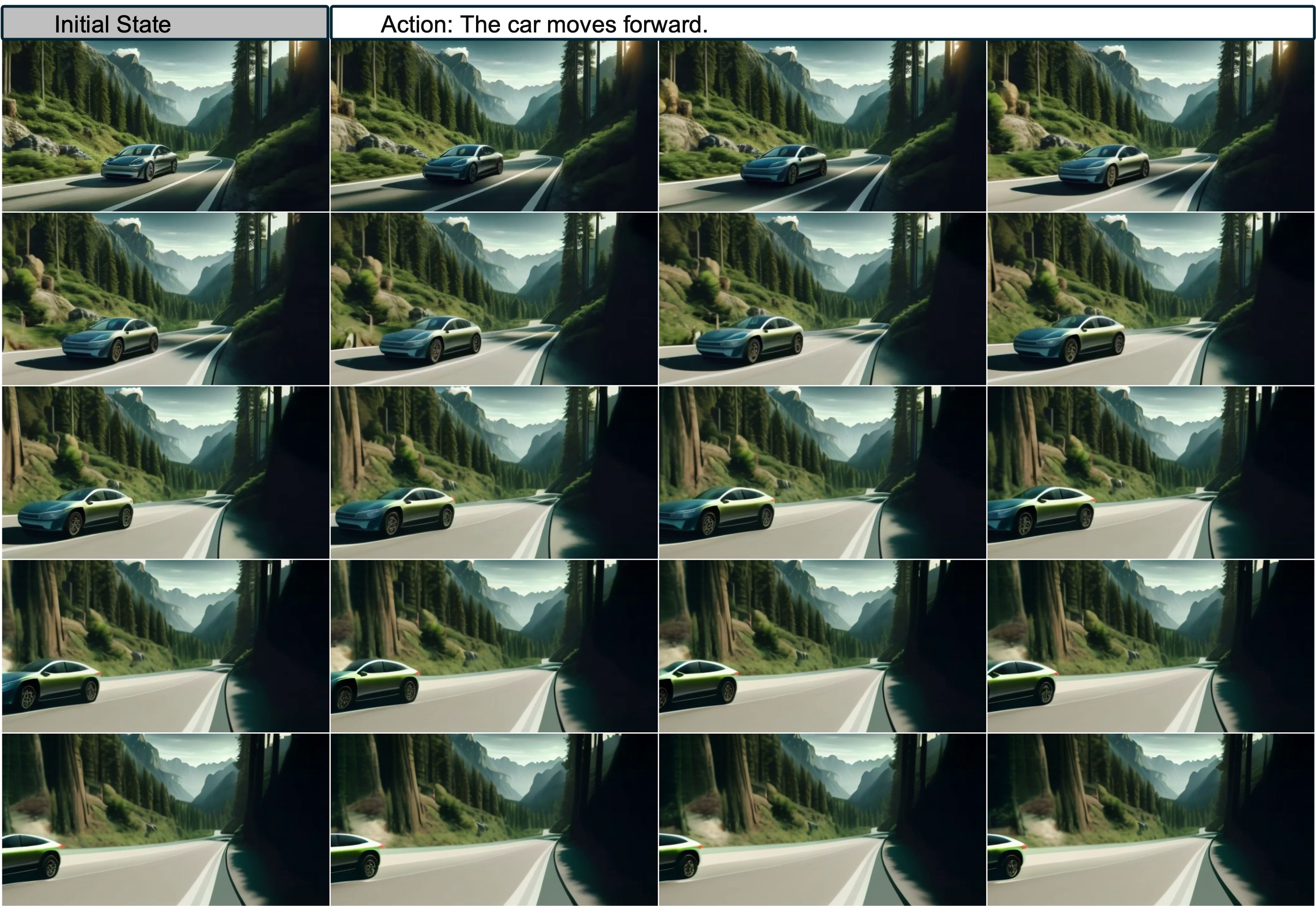}
    \end{subfigure}
    
    \begin{subfigure}[b]{1\textwidth}
        \centering
        \includegraphics[width=\textwidth]{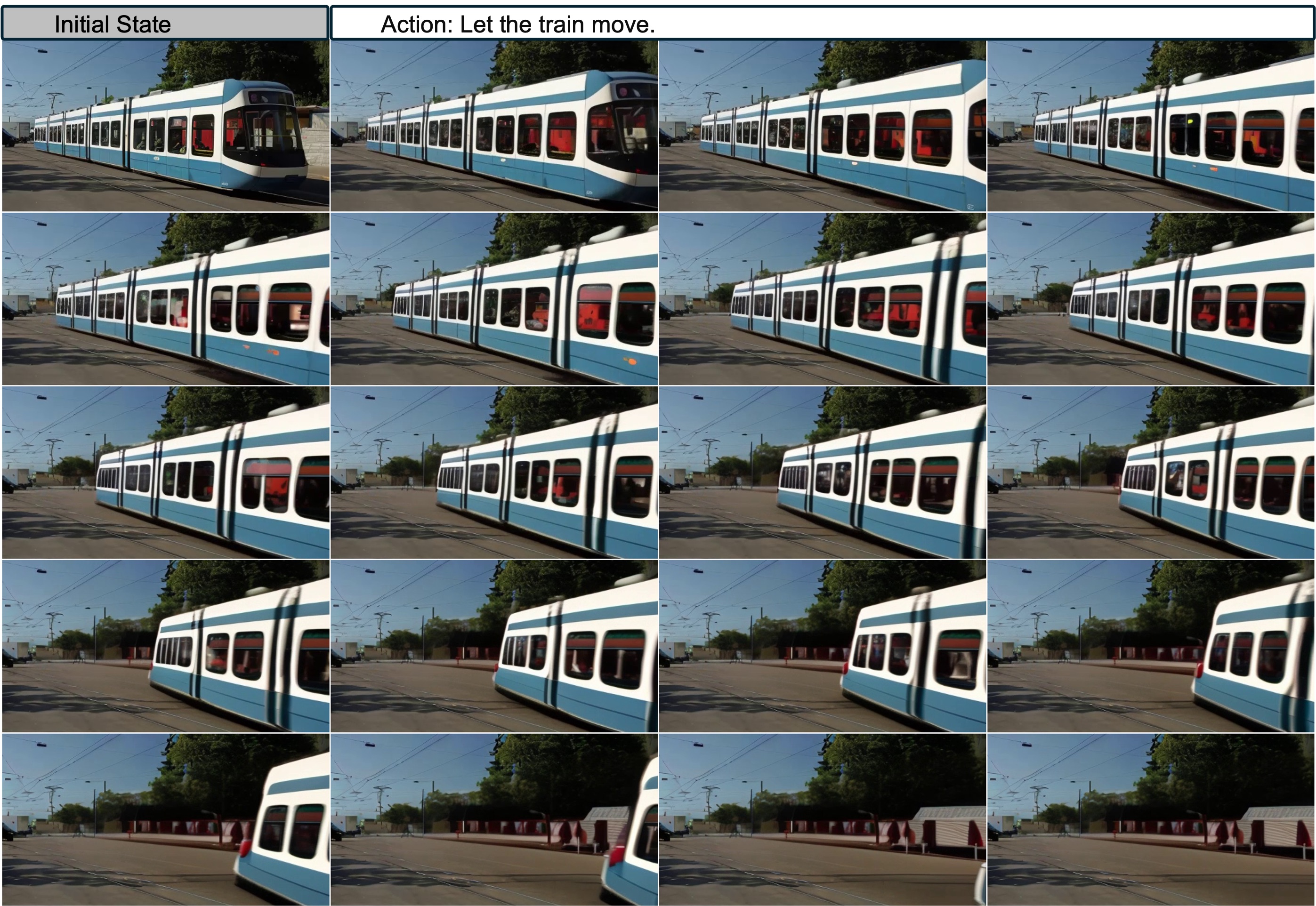}
    \end{subfigure}
    
    \caption{\modelname is capable of generating longer video autoregressively.}
    \label{fig:long-video}
\end{figure}

With the autoregressive backbone, \modelname is capable of generating longer videos of higher quality in an autoregressive manner. \modelname is trained on videos with up to 5 seconds (40 frames), but it is able to generate longer videos. We show the results of generating 8-second (64-frame) videos in Figure~\ref{fig:long-video}.

\clearpage

\subsection{Limitations} 
\begin{figure}[th]
    \centering
    \begin{subfigure}[b]{1\textwidth}
        \centering
        \includegraphics[width=\textwidth]{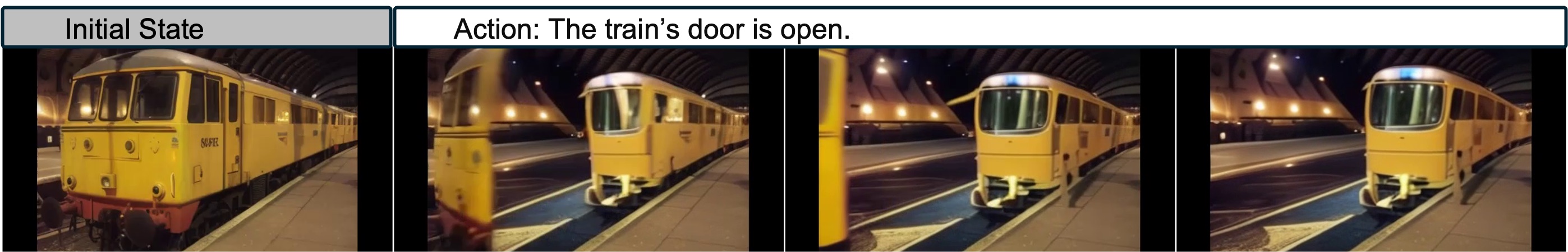}
    \end{subfigure}
    
    \begin{subfigure}[b]{1\textwidth}
        \centering
        \includegraphics[width=\textwidth]{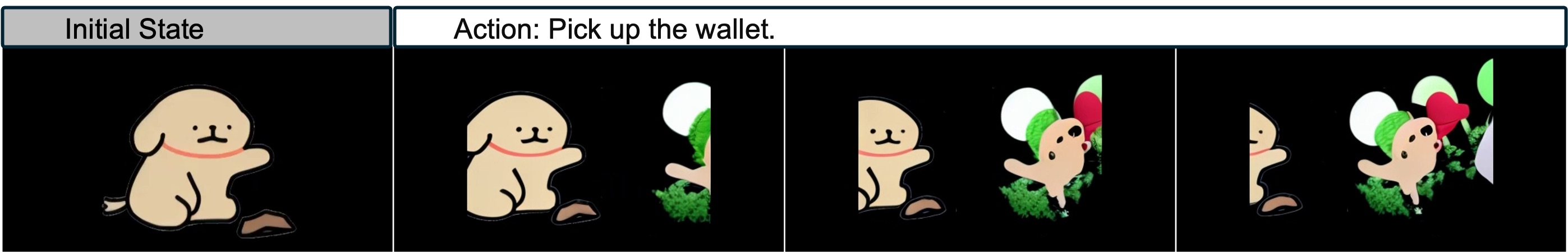}
    \end{subfigure}
    \begin{subfigure}[b]{1\textwidth}
        \centering
        \includegraphics[width=\textwidth]{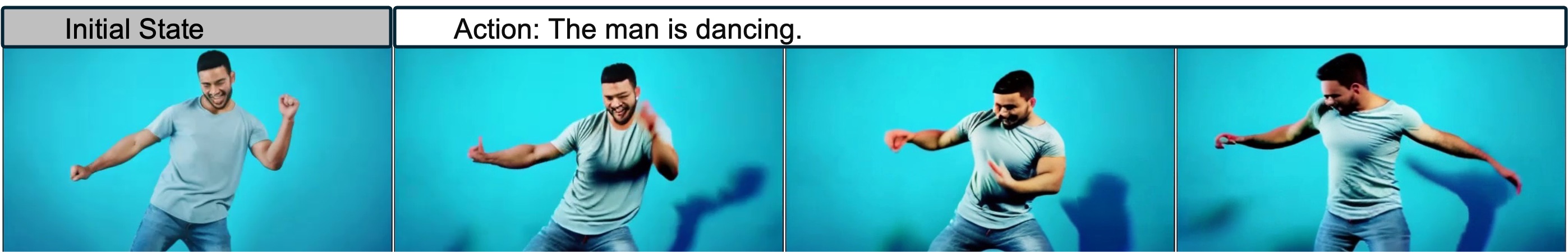}
    \end{subfigure}
    
    \begin{subfigure}[b]{1\textwidth}
        \centering
        \includegraphics[width=\textwidth]{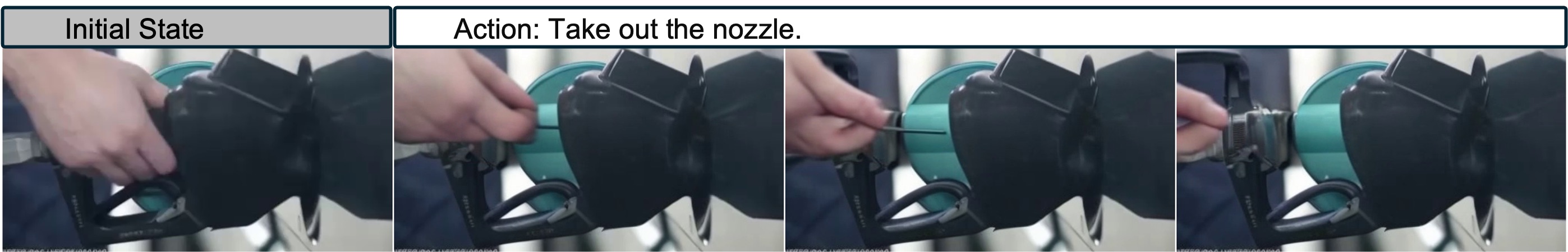}
    \end{subfigure}
    
    \caption{\modelname can fail in generating consistent videos, simulating complex scenarios, understanding commonsense and physical laws, and following instructions/actions.}
    \label{fig:failure}
\end{figure}

\modelname can struggle to generate videos with high quality and good controllability. Figure~\ref{fig:failure} shows failure cases about semantics understanding, motion control, and video consistency.

\begin{figure}[ht]
    \centering
    \begin{subfigure}[b]{1\textwidth}
        \centering
        \includegraphics[width=\textwidth]{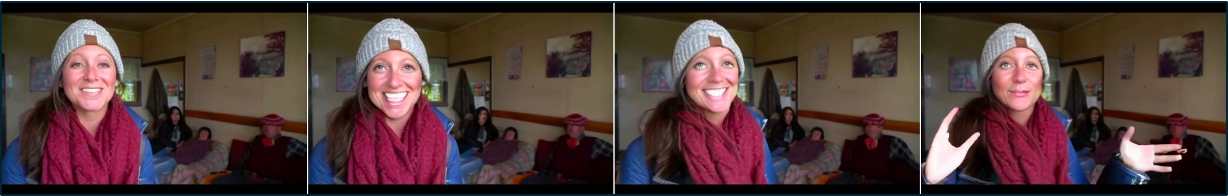}
         \caption{Results with small-scale training compute.}
    \end{subfigure}
    \vspace{0.2pt}
    
    \begin{subfigure}[b]{1\textwidth}
        \centering
        \includegraphics[width=\textwidth]{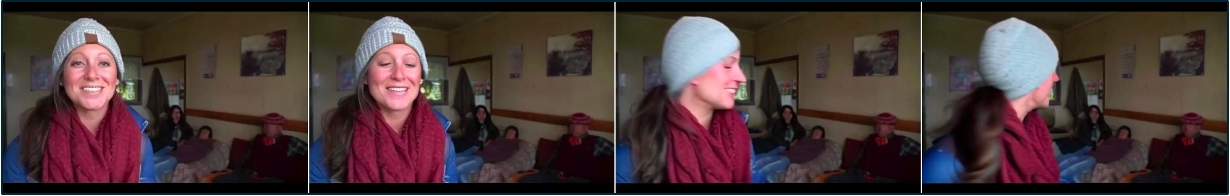}
         \caption{Results with large-scale training compute.}
    \end{subfigure}
    
    \caption{Action: "the woman turns her head left". After scaling-up, the model shows better video quality and controllability.}
    \label{fig:compute}
\end{figure}

When conducting small-scale exploratory experiments, we found that the data quality, \emph{i.e.}, the precision of the dynamics descriptions, has great influence on the model performance. In the domains where high-quality simulation data exists, the model easily gains great controllability. But in the domains of public video datasets, where captions generated by GPT-4 Turbo are noisy, the model does not show good performance. However, when we increased the training compute, controllability across general domains emergents on the model. We show a result comparison between the models trained with small-scale and large-scale training compute in Figure~\ref{fig:compute}. We hypothesize it is because increasing data size can mitigate some of the noises in the data. The results indicate the great potential of building a stronger general world model by larger-scale training.

\section{Related Works}

\paragraph{World models}
World models simulate the future state of the world based on its previous states and given actions \citep{tolman1948cognitive, briscoe2011mental, battaglia2013simulation, allen2020rapid, pramod2020evidence}. 
Previous world models in AI systems are usually designed for specific domains. For example, in robotics domain, world models are usually used for model-based reinforcement learning in specific simulators~\cite{recurrent-world-model,world-model,matsuo2022deep,chen2021transdreamer,kaiser2019model}. 
In robotics domain, world models~\cite{yang2023learning,zhen20243d,zhou2024robodreamer,ajay2023compositional,du2023learning} are capable of predicting future image or video states across diverse robotics environments. These predictive capabilities are important for robots to understand the environments, make informed decisions, and execute tasks accurately. 
Besides the robotics domain, world models are also widely used in autonomous driving~\cite{wang2023driving,wang2023drivedreamer,hu2023gaia,li2023drivingdiffusion,zhao2024drivedreamer,zhang2023learning,zheng2023occworld}, where they mainly focus on path planning and real-time decision-making, which is pivotal in enabling vehicles to navigate complex environments safely and efficiently.
There are also world models for 2D games~\cite{bamford2020neural,chiappa2016recurrent,eslami2018neural,hafner2019dream,kim2020learning,micheli2022transformers,robine2022transformer}. For example, Genie~\cite{bruce2024genie} is a generative model capable of simulating an interactive 2D game given an image. In this work, we make a step towards building a more general world model that simulates any-domain states given any-text actions at any time.



\paragraph{Video generation models}
Video generation models aim to synthesize realistic videos given text prompts or initial frames. 
%
Recent successes in diffusion models~\cite{ho2020denoising, rombach2022high,sohl2015deep, song2020score} have paved the way for their application in the video generation domain~\cite{khachatryan2023text2video, peng2023conditionvideo, zhang2023controlvideo, chen2023control, harvey2022flexible, singer2022make, tang2024any, voleti2022mcvd, wang2024magicvideo}.
For example, additional modules are introduced into the existing image diffusion models~\cite{ho2022video,ho2022imagen,blattmann2023stable,xing2023dynamicrafter,zhang2023i2vgen,zhang2024moonshot} to facilitate video generation capabilities.
However, the length of generated videos is limited due to the non-autoregressive nature.
Consequently, the Diffusion Transformer (DiT)~\cite{peebles2023scalable} has been proposed to allow for autoregressive generation, and Sora~\cite{videoworldsimulators2024} has further scaled it up, achieving remarkable success in generating long, high-quality video.
Furthermore, as the strong understanding and generation ability of LLMs, \cite{chameleonteam2024chameleon,yu2023scaling} have explored the usage of LLMs in vision generation domain. Additionally, \cite{kondratyuk2024videopoet,sun2024generative} incorporate LLMs for video generation to enhance the semantic understanding. Previous models are designed to generate scenes from input descriptions, yet they frequently lack the ability to control actions or predict real-world states. On the contrary, \modelname is a hybrid autoregressive-diffusion model, thus it is capable of on-the-fly control over video generation.

\section{Conclusion}
We presented \modelname as a step towards building a general world model. The model is able to simulate world states by generating videos across different domains, and control the video on the fly with natural language actions. \modelname introduces a staged training recipe that allows to reuse and integrate existing pretrained language  and video models. We believe larger-scale training with larger backbone models (e.g., GPT-4 and Sora) will lead to further improvement in terms of domain generality, video consistency, and action controllability. We are also excited about extending the model by incorporating other modalities, such as audio, to better measure and simulate the world. 


\clearpage


\bibliography{neurips_2023}
\bibliographystyle{plain}


\end{document}